\def\paperID{0000}
\def\confName{CVPR}
\def\confYear{2024}

\def\authorBlock{
Chao He,~~Jianqiang Ren,~~Jianjing Xiang,~~Xiejie Shen  \\ 

   Tongyi Lab,~~Alibaba Group\\
   
 {\tt\small \{yichao.hc, jianqiang.rjq, jianjing.xjj, shenxiejie.sxj \}@alibaba-inc.com}
}

\newif\ifreview 
\newif\ifarxiv \newcommand{\arxiv}{\arxivtrue}
\newif\ifcamera 
\newif\ifrebuttal 

\arxiv  

\documentclass[10pt,onecolumn,letterpaper]{article}
\ifreview \usepackage[review]{cvpr} \fi
\ifarxiv \usepackage[pagenumbers]{cvpr} \fi
\ifrebuttal \usepackage[rebuttal]{cvpr} \fi
\ifcamera \usepackage{cvpr} \fi

\usepackage{graphicx}	
\usepackage{amsmath}	
\usepackage{amssymb}	
\usepackage{booktabs}
\usepackage{times}
\usepackage{microtype}
\usepackage{epsfig}
\usepackage[table,xcdraw,dvipsnames]{xcolor}
\usepackage{caption}
\usepackage{float}
\usepackage{placeins}
\usepackage{color, colortbl}
\usepackage{stfloats}
\usepackage{enumitem}
\usepackage{tabularx}
\usepackage{xstring}
\usepackage{multirow}
\usepackage{xspace}
\usepackage{url}
\usepackage{subcaption}
\usepackage{xcolor}
\usepackage[hang,flushmargin]{footmisc}

\ifcamera \usepackage[accsupp]{axessibility} \fi

% ...

\ifarxiv  \fi

\newcommand{\R}[1]{{%
    \textbf{%
        \ifstrequal{#1}{1}{\textcolor{red}{R#1}}{%
        \ifstrequal{#1}{2}{\textcolor{blue}{R#1}}{%
        \ifstrequal{#1}{3}{\textcolor{magenta}{R#1}}{%
        \ifstrequal{#1}{4}{\textcolor{teal}{R#1}}{%
                           \textcolor{cyan}{R#1}%
        }}}}%
    }%
}}

\usepackage{xr-hyper}

\makeatletter
\newcommand*{\addFileDependency}[1]{
  \typeout{(#1)}
  \@addtofilelist{#1}
  \IfFileExists{#1}{}{\typeout{No file #1.}}
}

\makeatother
\newcommand*{\myexternaldocument}[1]{
    \externaldocument{#1}
    \addFileDependency{#1.tex}
    \addFileDependency{#1.aux}
}

\definecolor{cvprblue}{rgb}{0.21,0.49,0.74}
\usepackage[pagebackref,breaklinks,colorlinks,citecolor=cvprblue]{hyperref}
\usepackage[capitalize]{cleveref}
\crefname{section}{Sec.}{Secs.}
\crefname{table}{Table}{Tables}
\crefname{figure}{Fig.}{Figs.}

\ifarxiv \crefname{appendix}{App.}{Apps.}
\else \crefname{appendix}{Suppl.}{Suppls.} \fi

\usepackage{indentfirst}
\unless\ifarxiv \myexternaldocument{_supplementary} \fi

\begin{document}
 
\title{CartoonAlive: Towards Expressive  Live2D Modeling from Single Portraits}

\author{\authorBlock}
\maketitle

\begin{figure*}[h]
\vspace{-20pt}
\centering
\includegraphics[width=0.98\textwidth]{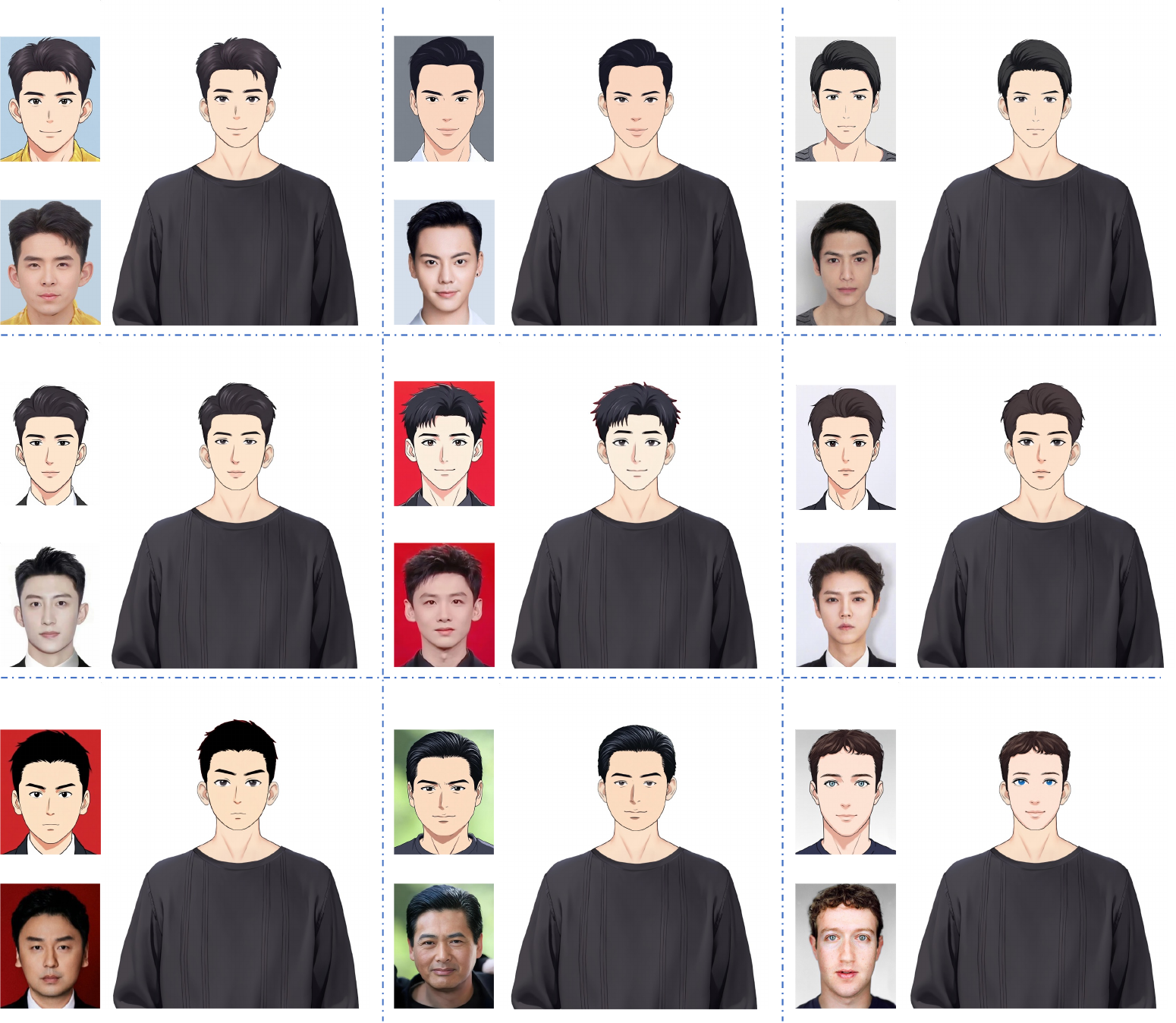}
\caption{Examples of animatable 2D cartoon characters generated by \textbf{CartoonAlive}. In each example, the bottom-left image shows a real human portrait, the top-left image displays a stylized cartoon version of the same subject, and the right side presents the resulting animated Live2D character model.}
\label{fig:live2d-results-9}
 
\end{figure*}

\begin{abstract}

With the rapid advancement of large foundation models, AIGC, cloud rendering, and real-time motion capture technologies, digital humans are now capable of achieving synchronized facial expressions and body movements, engaging in intelligent dialogues driven by natural language, and enabling the fast creation of personalized avatars. While current mainstream approaches to digital humans primarily focus on 3D models and 2D video-based representations, interactive 2D cartoon-style digital humans have received relatively less attention. Compared to 3D digital humans that require complex modeling and high rendering costs, and 2D video-based solutions that lack flexibility and real-time interactivity, 2D cartoon-style Live2D models offer a more efficient and expressive alternative. By simulating 3D-like motion through layered segmentation without the need for traditional 3D modeling, Live2D enables dynamic and real-time manipulation. In this technical report, we present \textbf{CartoonAlive}, an innovative method for generating high-quality Live2D digital humans from a single input portrait image. \textbf{CartoonAlive} leverages the shape basis concept commonly used in 3D face modeling to construct facial blendshapes suitable for Live2D. It then infers the corresponding blendshape weights based on facial keypoints detected from the input image. This approach allows for the rapid generation of a highly expressive and visually accurate Live2D model that closely resembles the input portrait, within less than half a minute. Our work provides a practical and scalable solution for creating interactive 2D cartoon characters, opening new possibilities in digital content creation and virtual character animation. The project homepage
is \url{https://human3daigc.github.io/CartoonAlive_webpage/}.

\end{abstract}

\section{Introduction}
\label{sec:intro}


Cartoon characters are widely used in films, games, social media, and advertising, typically appearing in either 2D or 3D forms. While 3D cartoon characters offer greater flexibility and realism, they often require high production costs and powerful rendering engines. On the other hand, traditional 2D video-based characters, although easier to produce, generally lack real-time interactivity and dynamic expressiveness.

Live2D \cite{live2d} technology bridges this gap by enabling interactive and expressive animation from static 2D illustrations. By simulating 3D-like deformation using layered 2D graphics, Live2D provides an efficient and cost-effective solution for creating animated characters that can be manipulated in real time. As a result, Live2D has become a leading standard for building expressive and interactive 2D digital humans, especially on platforms with limited computational resources.


In the real world, human facial appearances exhibit remarkable diversity—variations in eye shape, eyebrow style, nose size, lip thickness, and facial contour all contribute to an individual's unique identity. In Live2D modeling, the facial region is typically composed of multiple fine-grained components arranged across different layers. A major challenge lies in how to generate a wide variety of facial identities using a finite set of pre-defined parts. To address this, we propose \textbf{CartoonAlive}, an automated system capable of generating highly expressive Live2D models from a single input portrait image. The core innovation of \textbf{CartoonAlive} lies in its compositional generation capability, which draws inspiration from the shape basis concept in 3D face reconstruction to design blendshapes suitable for Live2D.


Specifically, given a single facial image as input, our method regresses the pose and scale parameters (horizontal shift $x$, vertical shift $y$, zoom-in and zoom-out parameter $scale$) of key facial components such as eyebrows, eyes, nose, and mouth. These parameters are learned through model training and optimized to ensure that the generated Live2D model closely matches the target identity. The entire process is fully automated and can generate a personalized Live2D character within 30 seconds, significantly improving efficiency and reducing production costs compared to traditional manual workflows or prompt-based generative approaches that rely heavily on user intervention.

The key features of \textbf{CartoonAlive} include:


\noindent {\bf Live2D Blendshape Design.}
We redesign the structure of Live2D models to support linear control of facial components along three axes: horizontal ($x$), vertical ($y$), and scaling ($scale$), with parameter ranges spanning from $-30$ to $30$. This enables the creation of a diverse range of facial expressions and identities. Additionally, we modify and expand the base face template to accommodate various facial types, including long, round, and broad faces.


\noindent {\bf Accurate Facial Parameter Prediction.}
To enable precise parameter estimation, we synthesize a large dataset of 100,000 paired samples by rendering 1024×1024 facial images at consistent positions using the PyGame rendering engine. For each rendered image, facial landmarks are extracted and matched with their corresponding Live2D parameters. We then train a Multilayer Perceptron (MLP) \cite{werbos1974beyond} to learn the mapping from facial landmarks to Live2D parameters. During inference, this network accurately predicts the necessary parameters based on the detected landmark positions from the input image.


\noindent {\bf Dynamic Artifact Correction.}
Once the facial parameters are obtained, the corresponding textures are placed accordingly. However, during animation, visual artifacts may occur due to misalignment between the foreground elements and the underlying face image; for example, when the eyes are closed, the background eyes may still be visible. To resolve this issue, we render facial masks based on the inferred parameters and use them to precisely identify the regions requiring inpainting. Guided by these masks, we repaint the underlying face image to eliminate visual inconsistencies, ensuring a dynamically flawless Live2D model during animation.


\noindent {\bf Hair Transfer.}
After aligning the facial contour, we perform hair segmentation on the input image to extract the hair mask, which is then transferred to the hair texture. If bangs occlude the eyebrows in the input image, we first remove the hair before extracting facial feature textures and parameters. Finally, we apply hair segmentation to the original image and transfer the hair to the final Live2D model.





Our main contributions in this work are as follows:

\begin{itemize}
\item To the best of our knowledge, CartoonAlive is the first fully automated framework that enables the end-to-end generation of complete Live2D cartoon characters from a single portrait image. Our method achieves this within 30 seconds, eliminating the need for further manual binding processes traditionally required in Live2D workflows.

\item Building upon the concept of shape bases in 3D face reconstruction, we introduce the idea of blendshapes for Live2D, and train a model to infer these parameters directly from facial features in the input image. Our approach allows the Live2D model to faithfully reproduce the facial identity of the input subject.

\item We propose a novel dynamic artifact correction mechanism that uses predicted facial parameters to render accurate facial masks. Guided by these masks, we repaint the underlying face image to eliminate visual artifacts during animation, resulting in a visually coherent and expressive Live2D model.
\end{itemize}

\section{Related Work}
\label{sec:related}

\begin{figure*}[tb]
\centering
\includegraphics[width=0.98\textwidth]{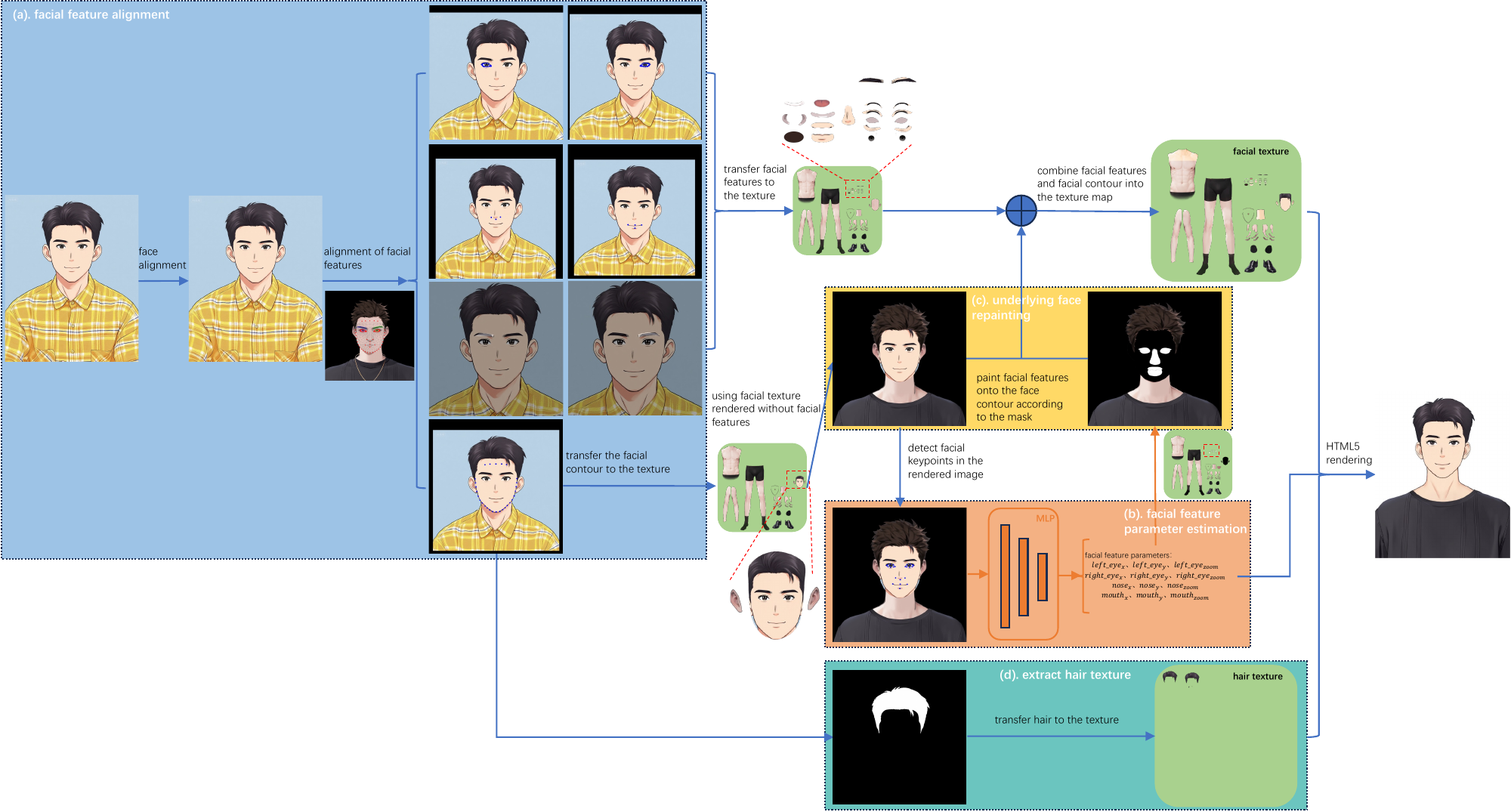}
\caption{Overview of the \textbf{CartoonAlive} pipeline. (a) \textbf{Facial Feature Alignment}: The input portrait is first preprocessed to align the eyes horizontally, ensuring consistent orientation. Then, facial keypoints for the eyes, nose, mouth, eyebrows, and facial contour are individually detected. A transformation is computed between each set of detected keypoints and those of a predefined template model. Based on this correspondence, each facial component in the input image is aligned accordingly.
(b) \textbf{Facial Feature Parameter Estimation}: Facial features are temporarily removed from the texture, and rendering is performed using only the underlying face image. Keypoints are then extracted from the rendered image, and corresponding Live2D parameters (e.g., position and scale) are inferred through a trained neural network.
(c) \textbf{Underlying Face Repainting}: To eliminate visual artifacts caused by overlapping facial features during animation, the underlying face image is repainted according to a mask derived from the inferred parameters, effectively removing foreground features that may interfere with dynamic expressions.
(d) \textbf{Hair Texture Extraction}: Hair segmentation is applied to isolate the hair region from the original image, which is then transferred into the final Live2D model as a separate texture layer. This ensures realistic integration of hair while preserving the integrity of facial components.}
\label{fig:pipeline-Imagelive}
\vspace{-5pt}
\end{figure*}

\noindent {\bf 3D Morphable Models (3DMM).}
The human face is a highly structured and complex object, whose composition can be effectively modeled using linear algebra principles—specifically, by representing each facial component as a basis vector in a high-dimensional space. Any individual face can then be expressed as a weighted combination of these basis vectors. In 1999, researchers at the Max Planck Institute introduced the 3D Morphable Model (3DMM) \cite{blanz2023morphable}, which defined a set of basis faces such that any given face could be reconstructed through a linear combination of these bases. This work laid the foundation for parametric modeling of facial geometry.

In 2009, the Basel Face Model (BFM) \cite{paysan20093d} was proposed, leveraging laser scanning data from 200 subjects to construct a detailed shape and texture model. BFM utilized Principal Component Analysis (PCA) \cite{abdi2010principal} to define a low-dimensional subspace that captures both facial shape and appearance variations. It was one of the first publicly available datasets and significantly advanced research in 3D face reconstruction. However, BFM's expression space remains relatively limited in capturing diverse facial expressions.

To address this limitation, the Max Planck Institute released FLAME \cite{li2017learning} in 2017, a more expressive and accurate open-source 3D face model built upon 33,000 facial scans. FLAME introduces separate shape, expression, and pose bases, allowing for fine-grained control over facial deformations. As one of the most widely used 3D face models today, FLAME has been instrumental in advancing facial reconstruction and animation tasks. These 3DMM-based methods provide a strong theoretical foundation for our approach, particularly in how we design and parameterize blendshapes in the 2D Live2D domain.

\noindent {\bf Text-to-Digital Human Generation.}
Recent advances in large language models (LLMs) have enabled the generation of digital humans based on textual descriptions. For example, Make-A-Character \cite{ren2023makeacharacter} utilizes LLMs to parse facial attributes from input text and employs Stable Diffusion \cite{podell2023sdxl} with ControlNet \cite{zhang2023adding} to generate reference images consistent with the described features. These images are designed to meet the requirements of subsequent 3D face fitting, ensuring frontal views and unobstructed facial components. The method ultimately produces 3D digital humans that accurately reflect the input description.

Our previous work, Textoon \cite{he2025textoon}, is the first to explore the generation of Live2D character models from text prompts. By decomposing the structure of Live2D models into modular components, Textoon enables fully automated creation of cartoon characters in under a minute without manual rigging. It supports rich variation in elements such as hairstyles, garments, and accessories, greatly expanding the diversity and expressiveness of generated characters. While Textoon demonstrates promising results in text-driven generation, it cannot support direct input from real-world portraits, which is the focus of our current work.

\noindent {\bf Image-to-Digital Human Generation.}
Reconstructing digital humans from a single image is a long-standing challenge in computer vision. Significant progress has been made in reconstructing 3D face geometry from monocular images, where deep learning has played a crucial role by reformulating the problem as a regression task. Existing approaches can broadly be categorized into two directions:

Coarse Shape Reconstruction: Early methods relied on annotated datasets to optimize 3DMM parameters, while recent learning-based approaches such as CNNs \cite{tewari2017mofa, deng2019accurate, danvevcek2022emoca, sanyal2019learning} and GCNs \cite{gao2020semi, lin2020towards} utilize large-scale in-the-wild datasets to learn robust representations. These methods reduce dependence on manual annotations and generalize well across diverse scenarios.

Fine-Grained Refinement: To capture detailed facial structures, displacement map-based methods \cite{lei2023hierarchical} have been proposed to refine coarse 3D reconstructions by adding residual deformations. These techniques enhance local geometry but often require additional supervision or assumptions about surface details.

Beyond 3D reconstruction, there has also been notable progress in generating 2D digital human videos from a single image. For instance, LivePortrait \cite{guo2024liveportrait} uses an implicit keypoint framework to extract facial features and contours from static images and drives dynamic expressions using motion information from external video inputs. Similarly, EMO \cite{tian2024emo} is an audio-driven talking head generation system that leverages diffusion models and attention mechanisms to synthesize expressive animations guided by voice input. While these works achieve impressive realism, they primarily target video-based outputs and do not directly support the creation of interactive 2D character models like those in Live2D.

\section{Live2D Generation}

This section provides a detailed description of our proposed method, \textbf{CartoonAlive}, for generating expressive Live2D models from a single input portrait. We begin by introducing the fundamental structure and principles of Live2D modeling, followed by an overview of our pipeline. Subsequently, we describe the core components of our system: facial feature alignment, blendshape parameter estimation, underlying face repainting, and hair texture extraction, as shown in Fig.~\ref{fig:pipeline-Imagelive}.

\subsection{Preliminary of Live2D}
\begin{figure}[tb]
    \centering
    \includegraphics[width=0.95\linewidth]{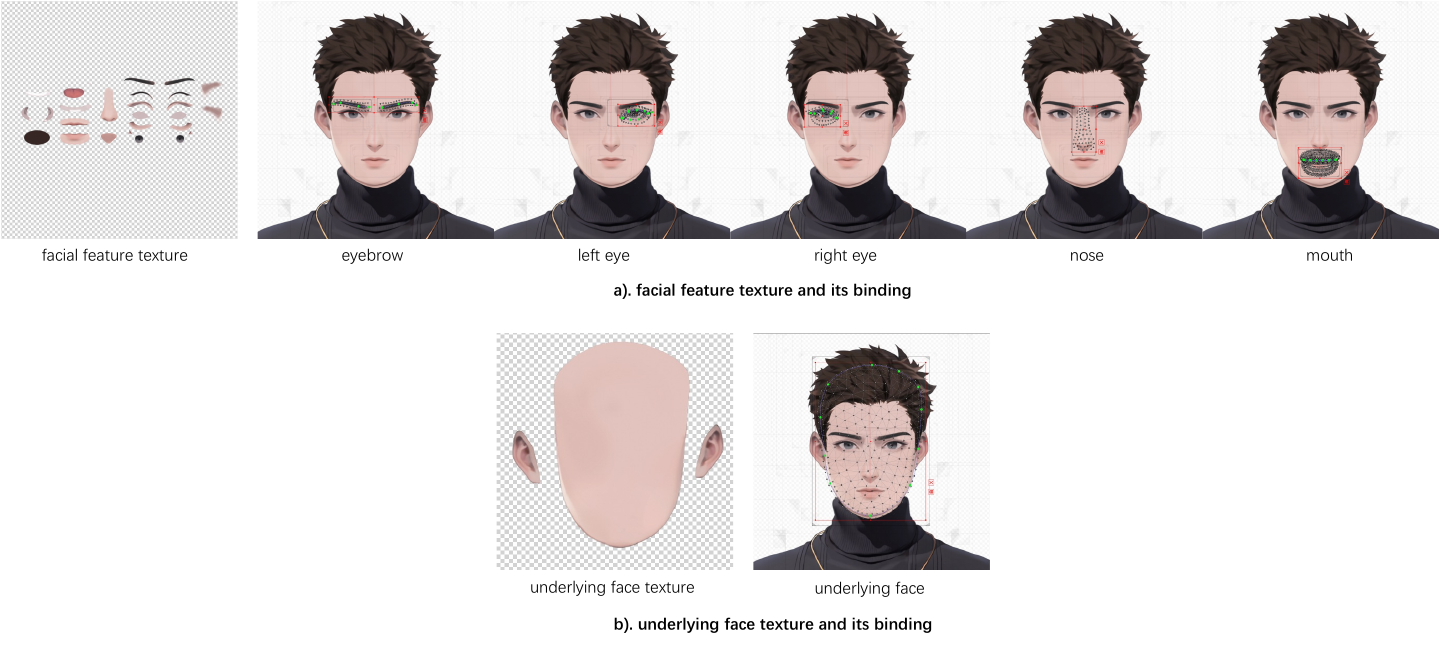}
    \caption{Live2D facial texture and its binding.}
    \vspace{-10pt}
    \label{fig:live2d-binding}
\end{figure}

A Live2D character is composed of multiple layered components, including the underlying face, eyebrows, eyes, nose, mouth, body, and clothing. Each component is defined by a polygonal mesh that controls its deformation during animation. As illustrated in Figure~\ref{fig:live2d-binding}(a), facial features such as the eyes, nose, and mouth are represented as separate layers within the texture map and bound to their respective meshes. In Figure~\ref{fig:live2d-binding}(b), the underlying face is also treated as a distinct layer, bound to an independent mesh. This hierarchical design ensures that dynamic changes to one part (e.g., closing the eyes) do not inadvertently reveal unwanted details from lower layers, thus preventing visual artifacts.

\subsection{Live2D Blendshape}
\begin{figure}[tb]
    \centering
    \includegraphics[width=0.6\linewidth]{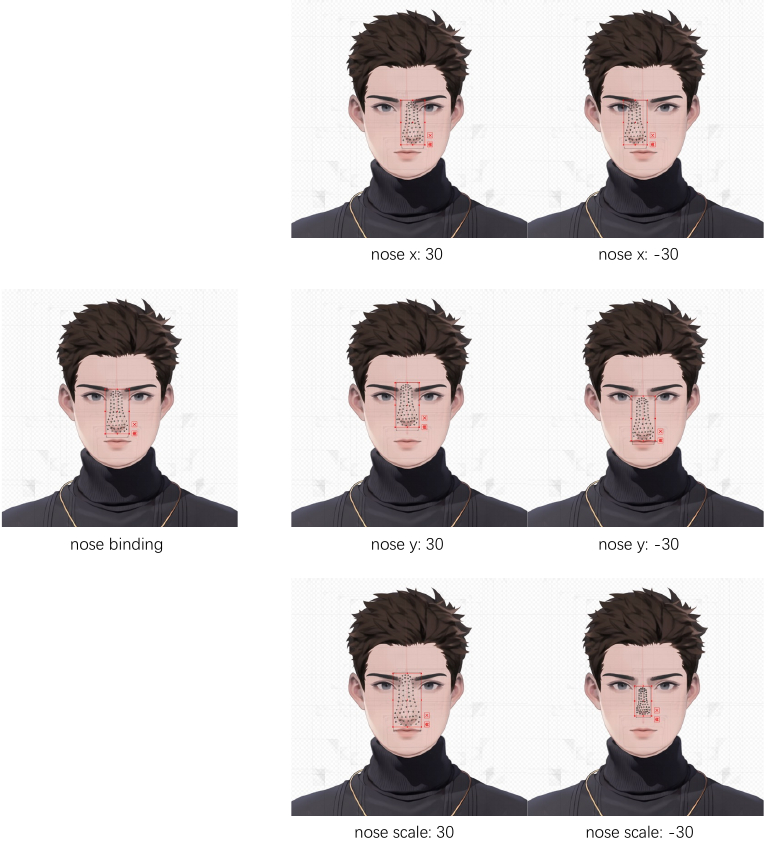}
    \caption{Nose blendshape creation with three dimensions: horizontal shift, vertical shift, and scale.}
    \label{fig:nose-blendshape}
\end{figure}

Inspired by the concept of shape bases used in 3D face reconstruction, we introduce \textbf{blendshapes} for Live2D modeling to enable flexible and identity-preserving facial deformations. For each facial component—such as the left eye, right eye, nose, and mouth—we define a set of basis shapes along three dimensions: horizontal shift ($x$), vertical shift ($y$), and scale. The weight coefficients for each dimension range from $-30$ to $30$, allowing fine-grained control over the position and size of each feature. An example of this approach applied to the nose is shown in Figure~\ref{fig:nose-blendshape}, where varying values across these dimensions generate diverse nasal appearances.

The overall facial configuration is modeled as a linear combination of these blendshapes:

$$
\mathbf{F} = \overline{\mathrm{F}} + \omega_{\text{left\_eye}} \mathbf{B}_{\text{left\_eye}} + \omega_{\text{right\_eye}} \mathbf{B}_{\text{right\_eye}} + \omega_{\text{nose}} \mathbf{B}_{\text{nose}} + \omega_{\text{mouth}} \mathbf{B}_{\text{mouth}}
$$

where $\overline{\mathrm{F}}$ denotes the base face with all parameters set to zero, and $\omega$ represents the blendshape weights. Each individual component can be further decomposed into its three-dimensional parameters, e.g.,

$$
\omega_{\text{left\_eye}} \mathbf{B}_{\text{left\_eye}} = \omega_{\text{left\_eye\_x}} \mathbf{B}_{\text{left\_eye\_x}} + \omega_{\text{left\_eye\_y}} \mathbf{B}_{\text{left\_eye\_y}} + \omega_{\text{left\_eye\_scale}} \mathbf{B}_{\text{left\_eye\_scale}}
$$

By varying these parameters, we can synthesize a wide variety of facial identities.

\subsection{Facial Feature Parameter Model Training}
\begin{figure}[tb]
    \centering
    \includegraphics[width=0.95\linewidth]{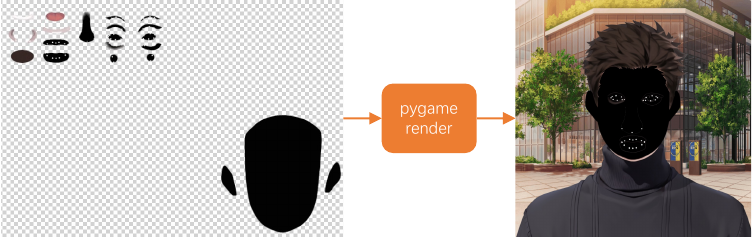}
    \caption{Accurate facial feature keypoints are obtained by detecting the white dots in the rendered images.}
    \vspace{-10pt}
    \label{fig:render-keypoints}
\end{figure}

To train a model capable of inferring Live2D blendshape parameters from facial landmarks, we construct a synthetic dataset based on the above formulation. Specifically, we randomly sample 100,000 sets of $\omega$ parameters and render corresponding facial images using a rendering engine. To ensure accurate landmark detection, we first black out the facial feature regions in the rendered images and mark the key points with white dots. These serve as ground-truth annotations for training. An example of the resulting facial keypoints is shown in Figure~\ref{fig:render-keypoints}.

Given the relatively low dimensionality of the parameter space, we use a 4-layer Multilayer Perceptron (MLP) for training. The input consists of normalized facial landmark coordinates, and the output corresponds to the predicted Live2D parameters. The network is trained using Mean Squared Error (MSE) loss until convergence.

\subsection{Facial Feature Alignment}
The input image $I$ is first aligned by rotating it so that the eyes are horizontally aligned, yielding $I_{\text{aligned}}$. Facial feature landmarks are then detected using Mediapipe~\cite{lugaresi2019mediapipe}. By matching these detected landmarks to a predefined template, we compute the transformation parameters required to align the input face with the template. The transformed image $I_{\text{transformed}}$ is then used to extract facial feature textures, which are mapped onto the Live2D texture map.

Furthermore, by computing the correspondence between the facial contour of $I_{\text{aligned}}$ and the template, we obtain the underlying face region and map it accordingly. This ensures that both the facial features and the underlying face are accurately positioned relative to the target model.

\subsection{Facial Feature Parameter Prediction}
With the transformed image $I_{\text{transformed\_contour}}$ aligned to the target model, we remove the facial feature components and render only the underlying face. Facial landmarks are extracted from this rendered image and fed into the trained MLP to predict the corresponding Live2D parameters $\omega$. Given the variability in eyebrow shapes and the potential inaccuracy in detecting eyebrow landmarks, we use a large bounding box to represent the eyebrow area, ensuring robustness in the final model.

\subsection{Underlying Face Repainting}
Although the predicted facial parameters $\omega$ allow us to reconstruct the static appearance of the input face, visual artifacts may occur during animation. For instance, when the eyes are closed, the underlying eyes might still be visible. To avoid this, we repaint the underlying face according to a mask derived from the inferred parameters. Specifically, we render a binary mask of the facial features and use it to erase the overlapping regions from the underlying face image, ensuring smooth transitions during animation.

\subsection{Hair Texture Extraction}
Hair segmentation is applied to extract the hair region from the input image. However, in cases where hair occludes facial features such as eyebrows or eyes, we employ a GAN-based hair removal model, HairMapper~\cite{wu2022hairmapper}, to clean the affected areas before performing facial alignment and parameter prediction. The final hair texture is extracted from the original image, preserving its natural appearance.

\subsection{Animation}
After the above pipeline, we are able to generate a static identity-consistent and dynamically artifact-free Live2D model. The resulting character not only preserves the appearance of the input face but also supports expressive animation driven by external controllers such as ARKit~\cite{arkitfaceblendshapes}. Compared to traditional Live2D models that typically rely on a limited set of parameters (e.g., MouthOpenY and MouthForm) for lip-syncing, our method leverages a more comprehensive parameter space consisting of 52 facial expression controls provided by ARKit. This allows for much richer and nuanced expressions, significantly enhancing the realism and interactivity of the animated characters.

\begin{figure}[tb]
    \centering
    \includegraphics[width=0.7\linewidth]{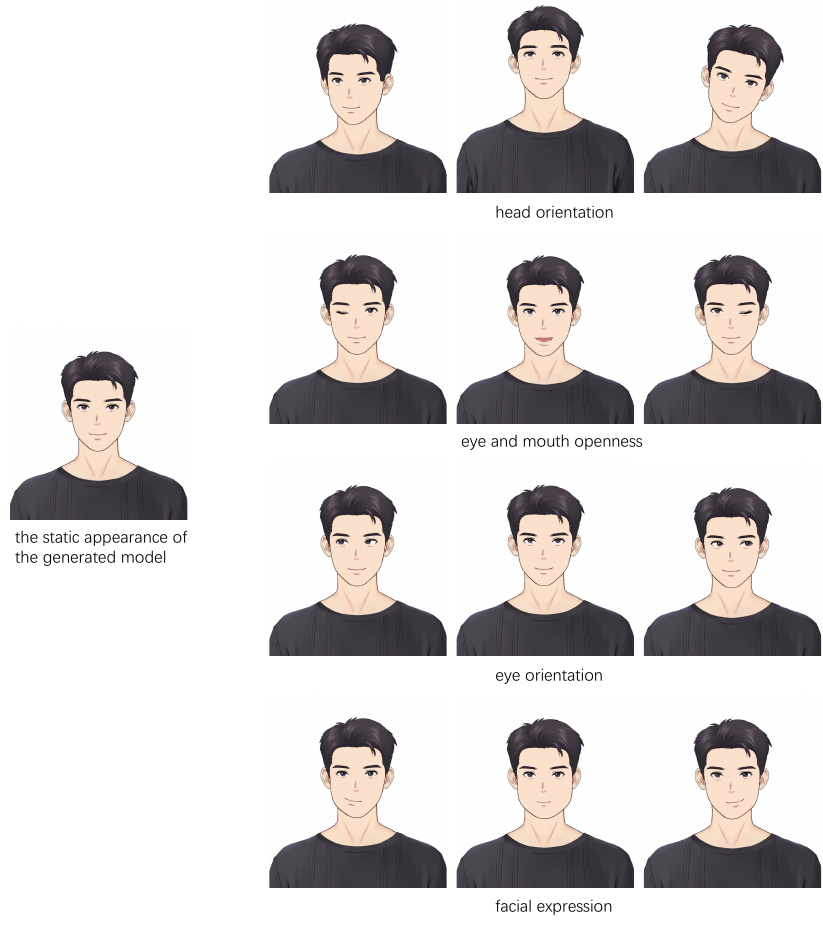}
    \caption{The overall animation effects of the generated Live2D model.}
    \vspace{-10pt}
    \label{fig:dynamic}
\end{figure}

Our model can be seamlessly integrated with existing animation pipelines and real-time interaction systems. As shown in Figure~\ref{fig:dynamic}, the generated Live2D character exhibits smooth transitions between different expressions and maintains high visual fidelity during animation. Furthermore, the dynamic behavior is consistent with the original input portrait, ensuring that the identity remains recognizable even under complex motion.

This level of expressiveness opens up new possibilities for applications in virtual communication, digital content creation, and AI-driven avatars, where both identity preservation and natural animation are essential.

\section{Results}
\label{sec:Results}

\begin{figure*}[tb]
\centering
\includegraphics[width=0.9\textwidth]{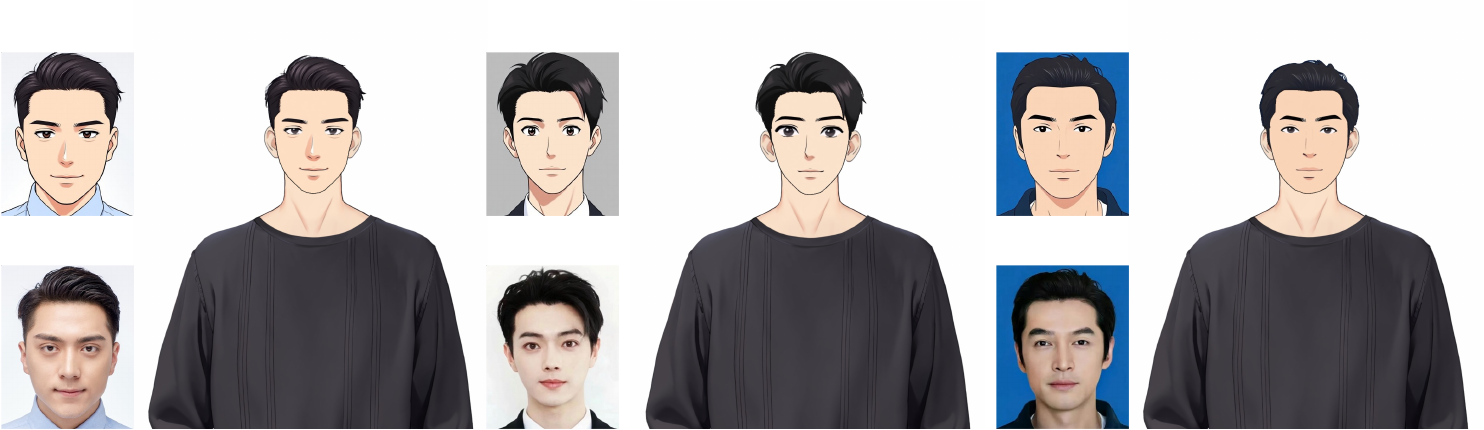}
\caption{Examples of Live2D cartoon characters created from input portraits.}
\label{fig:live2d-results-3}
\vspace{-15pt}
\end{figure*}

By integrating the aforementioned modules, \textbf{CartoonAlive} is able to generate a fully animated Live2D character from a single input portrait in less than 30 seconds. The results demonstrate high fidelity in preserving the identity of the input face while enabling smooth and expressive animations. A selection of generated characters is presented in Figure~\ref{fig:live2d-results-9} and Figure~\ref{fig:live2d-results-3}, which shows the effectiveness of our method in terms of visual quality and dynamic behavior.

\begin{figure*}[tb]
\centering
\includegraphics[width=0.8\textwidth]{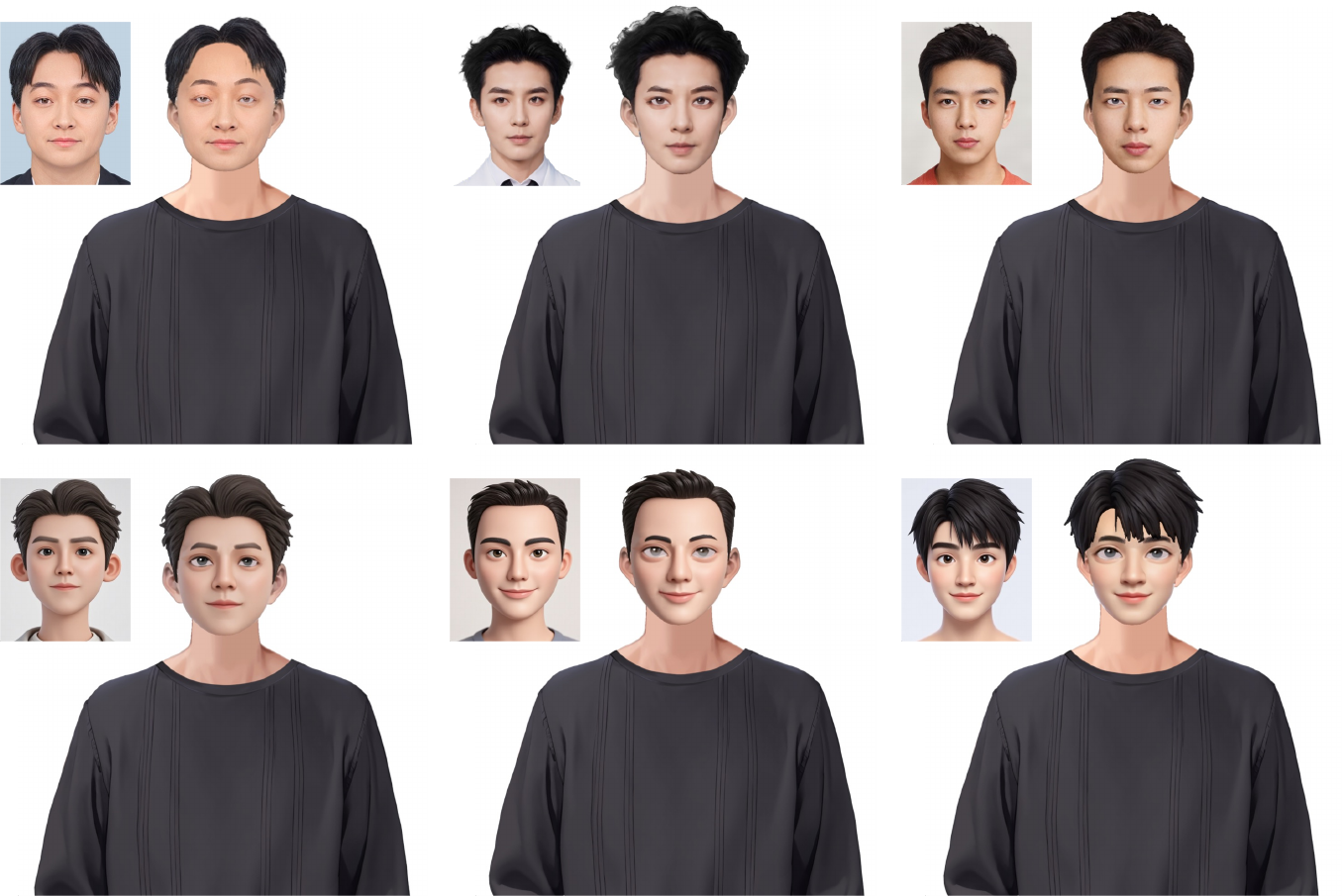}
\caption{Generated Live2D models from different artistic styles, including realistic human faces and 3D cartoon-style portraits.}
\label{fig:more_style}
\vspace{-15pt}
\end{figure*}

In addition to cartoon-style characters, we have also explored the application of our method on other artistic styles, such as realistic human faces and 3D cartoon-like portraits, as shown in Figure~\ref{fig:more_style}. While these variations present promising results, they also introduce additional challenges due to the increased complexity and fine-grained details involved.

\section{Limitation}
\label{sec:Limitation}
Despite the promising performance of \textbf{CartoonAlive}, there remain several limitations. First, due to the lack of reliable ear keypoint detection and the small number of pixels representing ear structures, the ears in the generated models are fixed and cannot match those in the input image. Second, the pupil and iris positions are often difficult to capture precisely due to their small size, leading to slight mismatches in eye appearance. Finally, although we utilize advanced segmentation techniques for hair, some fine strands may still be challenging to isolate, potentially affecting the accuracy of the final hairstyle representation.

\section{Conclusion}
\label{sec:Conclusion}
In this work, we present \textbf{CartoonAlive}, an efficient and fully automated method for generating identity-consistent, animatable Live2D characters from a single input portrait. By introducing the concept of blendshapes into the Live2D domain and leveraging a data-driven mapping between facial landmarks and model parameters, our approach enables rapid and accurate generation of personalized 2D avatars without manual intervention. Experimental results show that the generated characters not only preserve the identity of the input image but also support rich and expressive animation.

While challenges remain in capturing fine-grained facial features such as ears and hair, \textbf{CartoonAlive} establishes a new baseline for automatic Live2D character generation. It opens up exciting possibilities for applications in digital entertainment, virtual communication, and AI-driven content creation. Future work will focus on improving the accuracy of facial feature reconstruction and expanding the expressiveness of the generated models.

{\small
\bibliographystyle{ieeenat_fullname}
\bibliography{11_references}
}


\end{document}